\documentclass[conference]{IEEEtran}
\IEEEoverridecommandlockouts
\usepackage{latexsym}
\usepackage{amssymb}
\usepackage{amsmath}
\usepackage{amsthm}
\usepackage{booktabs}
\usepackage{enumitem}
\usepackage{graphicx}
\usepackage{color}
\usepackage{csquotes}
\def\BibTeX{{\rm B\kern-.05em{\sc i\kern-.025em b}\kern-.08em
    T\kern-.1667em\lower.7ex\hbox{E}\kern-.125emX}}

\begin{document}

\title{Understanding XAI Through the Philosopher's Lens: A Historical Perspective}

\author{\IEEEauthorblockN{1\textsuperscript{st} Martina Mattioli}
\IEEEauthorblockA{\textit{Ca' Foscari University of Venice} \\
Venice, Italy \\
martina.mattioli@unive.it}
\and
\IEEEauthorblockN{2\textsuperscript{nd} Antonio Emanuele Cinà}
\IEEEauthorblockA{\textit{University of Genoa} \\
Genova, Italy \\
antonio.cina@unige.it}
\and
\IEEEauthorblockN{3\textsuperscript{rd} Marcello Pelillo}
\IEEEauthorblockA{\textit{Ca' Foscari University of Venice} \\
Venice, Italy \\
pelillo@unive.it}
}

\maketitle

\begin{abstract}
Despite explainable AI (XAI) has recently become a hot topic and several different approaches have been developed, there is still a widespread belief that it lacks a convincing unifying foundation. On the other hand, over the past centuries, the very concept of explanation has been the subject of extensive philosophical analysis in an attempt to address the fundamental question of ``why'' in the context of scientific law. However, this discussion has rarely been connected with XAI. This paper tries to fill in this gap and aims to explore the concept of explanation in AI through an epistemological lens. By comparing the historical development of both the philosophy of science and AI, an intriguing picture emerges. Specifically, we show that a gradual progression has independently occurred in both domains from logical-deductive to statistical models of explanation, thereby experiencing in both cases a paradigm shift from deterministic to nondeterministic and probabilistic causality.
Interestingly, we also notice that similar concepts have independently emerged in both realms such as, for example, the relation between explanation and understanding and the importance of pragmatic factors. 
Our study aims to be the first step towards understanding the philosophical underpinnings of the notion of explanation in AI, and we hope that our findings will shed some fresh light on the elusive nature of XAI.
\end{abstract}


\section{Introduction}
Artificial Intelligence (AI) is becoming progressively a pervasive technology in our daily lives as a result of its increasing accuracy and versatility~\cite{Guidotti2019Survey}. 
Despite that, the growing integration of AI into human lives has determined a rising urgency to enlighten some of its potential undesirable outcomes. 
Consequently, its employment, particularly in contexts with paramount ethical considerations~\cite{ali2023explainable}, has led to the necessity for a fair decision-making process~\cite{lipton2018mythos,Ohara20}. 
These reflections have determined a variety of discourses about people's right to have an explanation of how the decision is reached by the machine, especially when the methods used are conceived as ``black boxes''~\cite{wachter2018counterfactual}. 
As a result of these considerations, scholars have posed various questions around, for example, when explanations are required, what models provide such explanations, what are the desiderata necessary to achieve understanding~\cite{doshi2017towards}, and what are the characteristics of a good explanation~\cite{ali2023explainable,miller2019explanation,Ohara20}. 
Within this debate, XAI is typically referred to as:
\begin{displayquote}
The process of elucidating or revealing the decision-making mechanisms of models. The user may see how inputs and outputs are mathematically interlinked. It relates to the ability to understand why AI models make their decisions~\cite{ali2023explainable}.
\end{displayquote} 
Nevertheless, defining explainability within the borders of a unique definition, amidst the plethora of those proposed, is a daunting task. 
Indeed, the majority of the aforementioned questions remain partially unresolved, to the extent that the precise definition of ``explanation'' remains to some degree obscure~\cite{doshi2017towards}.
Specifically, some authors contend that the ongoing discussion on explainable AI lacks a well-defined theoretical goal~\cite{Paez19pragmatic}. 
They argue that the concept of explanation, along with its related notions (e.g., interpretability~\cite{lipton2018mythos}), is ambiguously defined, thus fostering the perception that there is no cohesive and convincing conceptual foundation~\cite{doshi2017towards,lipton2018mythos}.
Additionally, it is worth noting that the variety of XAI models proposed is constantly evolving, which underscores the dynamic nature of this field~\cite{ali2023explainable} and its non-monolithic character~\cite{lipton2018mythos}. 
Indeed, abundant recent attempts have been made to classify and systematize these models (refer to~\cite{ali2023explainable,dwivedi2023explainable,Guidotti2019Survey} for in-depth surveys), reflecting a growing interest in XAI and the need for a more structured approach to its development~\cite{ali2023explainable}.

Despite this, the discourse surrounding explainability is not novel and has been explored in various contexts~\cite{Salmon1990}.  
Shifting our attention to the different realms of epistemology, this paper shows that analogous debates or inquiries arise. 
Indeed, the study of explanation has been a focal point of extensive philosophical analyses, undertaken to systematically address the fundamental question of ``why'' in the context of scientific law, thus unveiling one of the most substantial chapters in the philosophy of science~\cite{Salmon1990}. 
This discussion has a remarkable history, and its roots extend back to the philosophy of Aristotle, who distinguished between two types of knowledge: ``knowledge that'' and ``knowledge why,'' to wit description and explanation~\cite{Salmon1990}. 
Additionally, this distinction has become increasingly systematized over the past century, with a growing emphasis in scholarly discourses on the delineation and the proposal of a vast number of explanation models~\cite{Salmon1990}.
 
Acknowledging the significance of the epistemological discourse and the substantial inputs from philosophers in this domain~\cite{Salmon1990}, this paper investigates parallels and establishes a ``bridge'' between the discourse on XAI and the scientific explanation from the historical perspective. 
The objective of this study is to provide an epistemological framework that can assist in reinterpreting the concept of explanation through the lens of philosophy. 
In other words, we intend to understand XAI through the instruments of this rich philosophical literature to shed light on explainability and its elusive nature. 
Our purpose is to take a first step towards a deeper understanding of the philosophical underpinnings of the notion of explanation in AI by examining the historical debate that has taken place over the past centuries. 
Therefore, we posit that the ongoing discourse surrounding XAI, as it has unfolded in recent years, can be conceptually aligned with facets of the epistemological debate, as we reported in Figure~\ref{fig:chrono}.

In pursuit of this, Section~\ref{sec:related} illustrates previous emerging cross-domain works. In Section~\ref{sec:roots}, we discuss the philosophical roots of explanation and the relationship among some AI fields, including Machine Learning (ML), and science. 
We do so to establish the parallelism between scientific explanation and XAI. 
Section~\ref{sec:history} presents the epistemological debate on explanation, starting from Aristotle and reaching up to contemporary discussions. 
Finally, in Section~\ref{sec:comparison}, we compare the two discourses and underscore their interconnections.

\section{Related Works}
\label{sec:related}
In this section, we focus on incipient interdisciplinary efforts that have been done to connect and analyze psychological, sociological, and philosophical aspects of explanations. 
However, it should be emphasized that, to the best of our knowledge, no attempt has yet been made in the literature to link systematically the debates on XAI and scientific explanation.

\paragraph{Previous Philosophical Contributions.} 
Pioneering work in establishing connections between philosophy and XAI has been conducted by Páez~\cite{Paez19pragmatic}. 
The author elucidates the relationship between understanding and explanation both in the realm of scientific explanation and XAI. 
Subsequently, McDonnell~\cite{mcdonnell2023philosophy} provides some lessons from philosophy to assess better explanations. 
More specifically, his three primary observations include the necessity of a contrastive structure, the importance of focusing on actionable interventions, and the idea that robust causal dependence enhances the effectiveness of an explanation.  
Durán~\cite{duran2021dissecting} claims that scientific explanations are furnished with a precise structure aimed at providing a comprehensive understanding of the world. 
Also, his paper asserts that current XAI models do not qualify as genuine explanations.  
Finally, O'Hara~\cite{Ohara20} clarifies the relationship between explanation and understanding, establishing a connection with the decision process.

\paragraph{Explanation and Social Aspects.} 
A segment of the present literature has directed its attention towards social attributes of explanations, linking them with XAI. 
Miller~\cite{miller2019explanation} affirms that insights about humanities can benefit XAI. 
He emphasizes that explanations are contrastive, social, and selected in a biased manner and also that causal relations are more influential than probabilities. 
Mueller et al.~\cite{mueller2021principles} claim that there is a necessity for human-inspired XAI guidelines, as psychological principles often remain underestimated.  
Hoffman et al.~\cite{hoffman2018metrics} assert that explanations are not properties of statements, but result from interactions. 
In fact, what qualifies as an explanation depends on the learner's needs, previous knowledge, and goals. 

\paragraph{The Call for Clarification.}
Several authors called for clarity, remarking on the need for greater rigor in the definition of explainability and related concepts.
Lipton~\cite{lipton2018mythos} considers the term~\textit{interpretability} as slippery and ill-defined. 
Páez~\cite{Paez19pragmatic} argues that explanatory strategies may lack a precisely defined theoretical purpose. 
Boge and Poznic~\cite{boge2021machine} affirm that discussions on the philosophy of science could benefit ML. 
They emphasize the significant connections between these two disciplines and assert that the development of XAI could become a crucial theme in the philosophy of science.


\section{Philosophical Roots of Explanation}
\label{sec:roots}

\begin{figure*}
        \centerline{\includegraphics[width=0.99\textwidth]{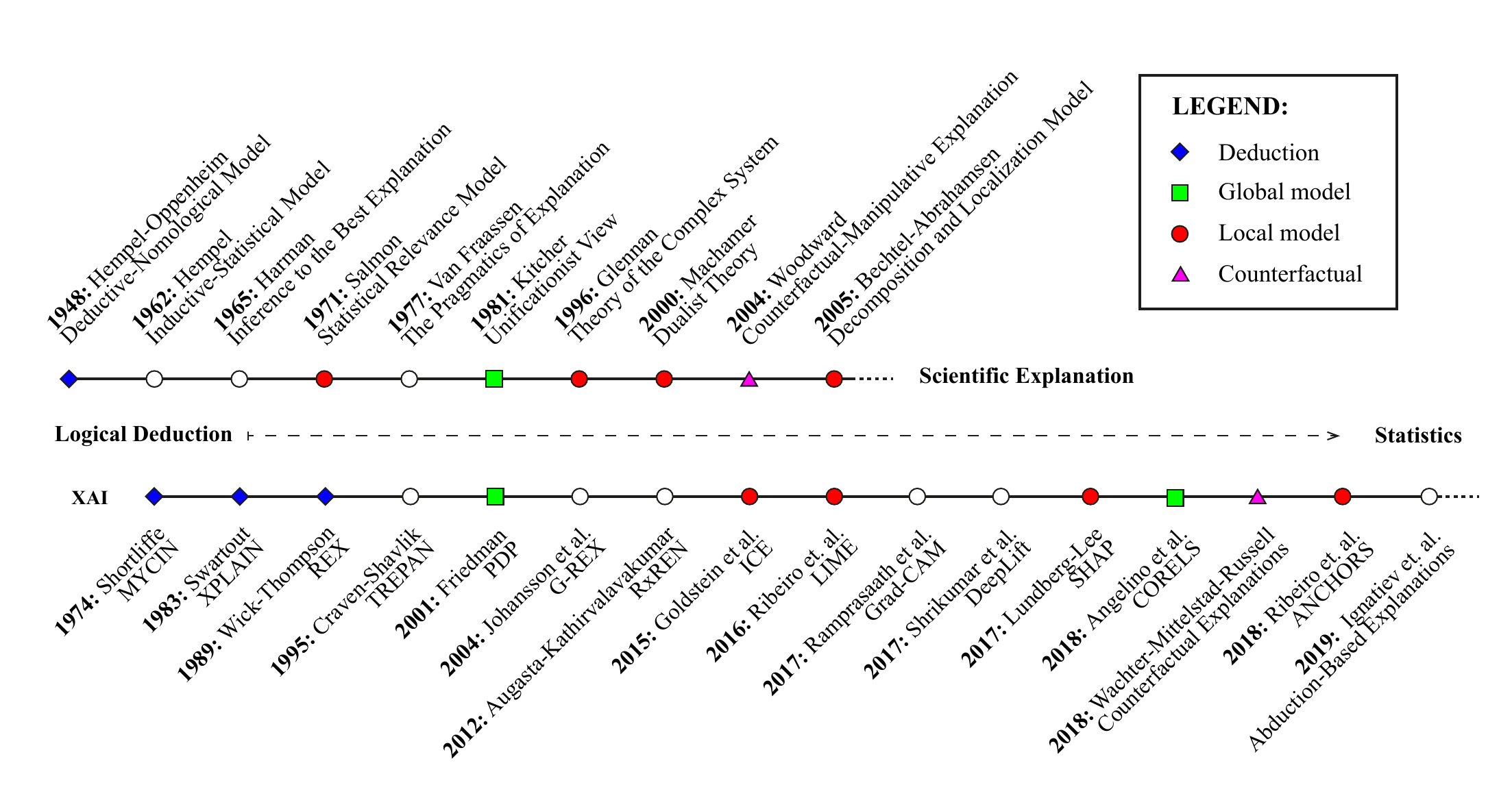}}
         \caption{Timeline of Scientific Explanation and XAI. The upper line represents the chronological development of the philosophical models, while the lower line illustrates the evolution of XAI. The middle line represents the general gradual change from deductive explanations to statistical ones. Analogies have been highlighted, as shown by the legend.}
         \label{fig:chrono}
    \end{figure*}

Embedded in centuries of philosophical inquiry, the fundamental concept of explanation has an extremely long tradition and ancient roots, which can shed light on the actual discussion on XAI. 
In particular, during the last decade, epistemology has been involved in a lively debate about scientific explanation, in which philosophers have meticulously delineated its constituents, seeking a precise definition while also reflecting upon the criteria of good explanations and discussing which particular model should be preferred to achieve them~\cite{Salmon1990}. 
As we aim to assert, the term ``explanation'' carries connotations and meanings that can be transposed to the current discussion regarding XAI, but that are beyond its common-sense definition or recent discussions, owing to the depth of the philosophical tradition in which it has been expounded.
Additionally, the relationship between scientific explanation and XAI is relevant not only because of potential parallels in the philosophical discourse or overlapping terminology in both debates. 
It also arises from a broader conception that originates from the proximity of some AI fields, including ML, with the scientific inquiry~\cite{Pelillo2013HowMI}. 
This closeness may contribute to extending and reinterpreting some of the implications of explanation, through a philosophy of science lens. 
Indeed, various authors provide valuable philosophical insights into these issues, contending that Pattern Recognition (PR) and ML are inherently aligned with scientific endeavors in their contribution~\cite{duin2007science,Pelillo2013HowMI}. 
This correspondence is evident in their pursuit to address similar questions related to categorization, causality, generalization, the problem of induction, or other pivotal aspects~\cite{duin2007science}. 
Similarly, the concept of explanation can benefit from a philosophical perspective.
However, before we delve into the depths of explainability, this section introduces some notions that may help elucidate the presentation of our parallelism within the context of the philosophy of science, including a brief presentation of the XAI debate, considerations regarding science as a ``black box,'' and pertinent philosophical terminology that will serve as a foundation to initially outline the concept of explanation.

\subsection{Short History of XAI}
Although the debate on explainability has recently gained prominence, particularly after the introduction of the right to explanation within the GDPR~\cite{ali2023explainable,wachter2018counterfactual}, this concept traces back to the early days of expert systems~\cite{confalonieri2021historical}. 
For instance, MYCIN~\cite{shortliffe1974rule} is a rule-based expert system, developed to help doctors select antimicrobial therapy. It includes a general question answerer and a status checker, enabling the physician to understand both the program's advice and its reasoning. 
This type of system is grounded in a hypothetico-deductive strategy and exhaustively applies inference rules~\cite{clancey_fg_nodate}, implying determinism~\cite{friedman_use_1983} and making the models easily interpretable~\cite{confalonieri2021historical}. 
REX~\cite{wick1989reconstructive} consists of a knowledge-based explanation system and a knowledge-based problem-solving system, in alignment with the existing epistemological separation between ``knowledge that'' and ``knowledge why.'' 
It offers explanations of how an expert system progresses from specific data to a final output. 
Differently from early AI systems, most ML models are not directly interpretable and can be considered as a ``black box''~\cite{confalonieri2021historical,lipton2018mythos}. 
Hence, explainability in this latter instance can be seen as finding a more interpretable surrogate model approximating the original one~\cite{ding2022explainability}. Consequently, the most popular XAI methods often lack rigorous guarantees~\cite{marques2023explainability}. As an alternative to heuristic or informal techniques~\cite{lundberg2017unified,ribeiro2016should}, growing interest has been posed on formal XAI, which offers logic-driven methods for deriving explanations, by providing theoretical assurances~\cite{Izza021}. 
Among these approaches, abductive explanation~\cite{Amogud21Abductive} stands out as an argument-based local explanation, consisting of a minimal set of literals sufficient for predicting a class. Thus, it serves as a reason for assigning a class to an instance~\cite{Amogud21Abductive}.
Moreover, runtime verification allows the explanation of AI-based self-adaptive systems, enabling the investigation of system behavior~\cite{havelund2015rule}. Finally, we cite XAI techniques built upon AI diagnosis principles, which involve identifying system faults or anomalies through logical reasoning and inference techniques~\cite{ignatiev2019model}.
However, the dichotomy between surrogates and formal explanations will be analyzed in Subsection~\ref{sec:deeper}, as crucial for the discussion in relation to~\textit{bona fide} explanations.

In general, due to the vastness of the discussion, several criteria are introduced to classify explainability in ML literature. 
For instance, a separation is established between global or local methods, depending on whether their goal is to explain the whole model or a single prediction. 
Also, there is a distinction between model-specific and model-agnostic approaches, relying on the fact that the explanation applies to a single model (or a group), or all ML ones~\cite{dwivedi2023explainable,Guidotti2019Survey}. 

Other salient taxonomies distinguish between feature-based or example-based techniques~\cite{dwivedi2023explainable} or between attribution, visualization, example-based, game theory, and knowledge extraction explanations~\cite{ali2023explainable}. 
Most of the relevant models identified in the pertinent literature have been reported in Figure~\ref{fig:chrono}, among them is worth mentioning the Counterfactual Explanation~\cite{wachter2018counterfactual}, a model-agnostic method that shows what change in features should be done to determine a prediction switch. 
Additionally, there exists LIME~\cite{ribeiro2016should}, which uses a linear classifier for a local approximation of the model to be explained.
Finally, we also mention SHAP~\cite{lundberg2017unified}, which links game theory to local explanations, by using the Shapley values. Specifically,  Shapley values assigns ``payouts'' to ``players'' based on their contributions to the ``total payout.''

\subsection{Black Box-ness Insights from Science}
The term ``black box'' is often employed to describe a model lacking interpretability, deemed antithetical to the principle of transparency, i.e., the property of an algorithm that is directly comprehensible~\cite{lipton2018mythos}. 
However, the metaphorical notion of a ``black box'' has received considerable attention in a wider range of disciplines, including but not limited to, science, philosophy of science, and psychology. 
Its interpretive significance extends beyond the field of ML and comprises a variety of theoretical frameworks and intellectual pursuits~\cite{bunge_1963,Hanson1963-HANTCO-27}. 

Hanson~\cite{Hanson1963-HANTCO-27}, for instance, introduced the concept of the ``black box'' as one of three stages in scientific development. Initially viewed as an algorithm with opaque internals, theories progress to a ``gray box'' stage where some structure is discernible, and finally to a ``glass box'' stage, offering transparent insights across disciplines. Additionally, when it comes to the ``black box'' nature of a model, the issue of explanation also arises. 
Within this whole theoretical framework, the term ``black box'' is employed as a metaphorical device to connote the idea that the system in question is, in some sense, a closed entity whose internal workings are inaccessible to outside scrutiny. 
Both AI and science can be interpreted within this definition~\cite{bunge_1963,Hanson1963-HANTCO-27,lipton2018mythos}. 

\subsection{Explanation Terminology Basics}
Before presenting the centuries-long philosophical dialogue, we provide some basic philosophical terminology and concepts. 
Quoting Salmon:
\begin{displayquote}
    Unless we take preliminary steps to give some understanding of the concept we are trying to explicate — the explicandum — any attempt to formulate an exact explication is apt to be wide of the mark~\cite{salmon1984scientific}.
\end{displayquote}
It is commonly accepted that science aims for knowledge acquisition about the world, distinguishing itself from common sense knowledge~\cite{Nagel1961-NAGTSO-3}. 
However, philosophical literature traditionally differentiates between two types of scientific knowledge, namely ``knowledge that'' and ``knowledge why.'' 
Indeed, the first concerns description, while the latter explanation~\cite{Salmon1990}. 
In particular, an explanation, which provides a scientific understanding of the world, is typically divided into two components: the ``\textit{explanandum}'' and the ``\textit{explanans}.'' 
The former pertains to the statements regarding the event requiring an explanation, whereas the latter encompasses those used to provide them~\cite{HempOpp48}. 
Another common concern relates to the nature of the phenomena requiring explanation, which can comprise individual events, general laws, or statistical regularities. 
According to Nagel~\cite{Nagel1961-NAGTSO-3}, there are four distinct explanation patterns since ``why questions'' are not all of the same type. 
These include deductive, probabilistic, functional or teleological, and genetic models of explanations. In deductive explanations, the ``explanandum'' is a logically necessary consequence of the explanatory premises. 
Probabilistic explanations stem from statistical premises, addressing individual cases. Functional explanations indicate the instrumental roles a unit has in bringing about a goal within a system. 
Lastly, genetic explanations delineate the sequence of significant events leading from an earlier system to a later one.


\section{The Historical Evolution of Scientific Explanation Debate}
\label{sec:history}
In this section, we aim to analyze the scientific explanation debate to acquire a comprehension of the issues and the philosophical foundations of explanation, providing useful insights into the multi-faced underpinnings of XAI discourse. 
Throughout the past discussions, a variety of positions have emerged within the epistemology framework, as well as analogous topics. 
For instance, consisting of the scarcity of accurate terminology or the challenges of selecting the optimal model for factoring in explanations~\cite{Salmon1990}. 
However, systematic attempts to solve these issues have been proposed in the epistemological literature, offering fruitful philosophical insights for XAI. 
To establish a correlation between two distinct debates and to identify potential intersections, we categorize the epistemological discussion into three distinct eras, in relation to Hempel and Oppenheim's turning point proposal of the Deductive-Nomological (D-N) model~\cite{HempOpp48}. 
These eras, namely the pre-Hempelian era, the received view, and the post-Hempelian era follow the chronological development. 
Our aim is, as illustrated in Figure~\ref{fig:chrono}, to highlight possible common trends and pivotal points of the discourses regarding the concerns raised.

\subsection{Pre-Hempelian Era}
Many of history's most eminent philosophers and scientists have questioned the nature of explanation and its role in science. 
However, it is not possible to answer by providing a unique definition. 
Instead, we should respond by starting from the very initial explorations. 
According to Aristotle~\cite{charlton1983aristotle}, it is only when we know the causes, or ``\textit{aitia},'' of something that we have an explanation for it, emphasizing the importance of explanation in response to ``why questions.''
Indeed,
\begin{displayquote}
   The discussion of aitia, on the other hand, is rather a discussion of explanation, and the doctrine of the ``four causes'' is an attempt to distinguish and classify different kinds of explanation, different explanatory roles a factor can play~\cite{charlton1983aristotle}.
\end{displayquote}
To be more specific, Aristotle identified four causes, which are different types of answers to the ``why question,'' namely the material cause, the formal cause, the efficient cause, and the final cause. 
In the Aristotelian view, causality and explanation are intimately related and, as we will see, causation assumes a key role in numerous accounts of explanation. 
However, not all philosophers have supported the notions of causality and explanation. 
For instance, in Galileo Galilei's various scripts, it is possible to recognize strong positions against the existence of causal relationships, to the extent that he affirmed that investigations on the causality of scientific phenomena are, not only worthless but also a fantasy~\cite{galilei1914dialogues}.
It becomes clear, as the debate unfolds, that the scientific community has not always unanimously accepted the idea of explanation as a distinct objective of science. 
Indeed, during the early positivist era, proponents of this school of thought categorically rejected the prospect of scientific explanation, seeking to counteract super-empirical influences originating from idealism. 
This refusal stemmed from the fact that many idealist philosophers' theories were instilled with transcendental metaphysics and referred to explanations involving extra-scientific factors~\cite{Salmon1990}. 
Consequently, this notion was, for an extended period, met with resistance in the discourse of the philosophy of science, being deemed an extraneous element beyond the scope of scientific inquiry. 
Therefore, the pursuit of answers to questions regarding causation, namely the ``why questions,'' was considered impossible or worthless~\cite{bunge2017causality}. 
This belief has been carried forward, for instance, by philosophers and scientists such as Mach~\cite{mach_science_2013} and Duhem~\cite{duhem_aim_1982}, who rejected the idea of evaluating physical theories based on their explanatory power, instead of their descriptive adequacy.

The paradigm shift occurred with logical empiricism, which began asserting that one of the purposes of science was the formulation of explications of fundamental concepts. 
Carnap~\cite{carnap1988meaning}, at the forefront, proposed his explanation view, distinguishing between two terms: the ``\textit{explicandum}'' and the ``\textit{explicatum}.'' 
The process of explication is the transformation from the ``explicandum'' to the ``explicatum'' and involves the conversion of an imprecise and pre-scientific concept into a new and precise one. 
Carnap's view provided the basis for the upcoming discussion on scientific explanation and the proposal of the ``received view,'' namely the Deductive-Nomological model~\cite{carnap1962logical}. 

\subsection{The Received View}
In 1948, the work of Hempel and Oppenheim brought the concept of explanation to the forefront of the philosophy of science, marking a pivotal moment in the trajectory of future debates, to the extent that it is possible to distinguish between the philosophical inquiry that happened before Hempel, and that that occurred after. 
While their model is often regarded as the first attempt to incorporate explanation into scientific discourse, their true contribution was to propose a structured effort at the systematization of scientific explanation into the so-called Deductive-Nomological model~\cite{HempOpp48}.
The core of their model lies in subsuming the ``explanandum'' under general laws and statements about the conditions under which the phenomenon occurred, through deductive inference. 
Accordingly, in a Hempelian context, to explain means to bring phenomena back into the realm of laws having empirical scope. 
An example would be helpful to have a better grasp of the Deductive-Nomological model. 
The ``explanandum'' consists of the description of the phenomenon to be explained, such as an oar underwater that appears bent upwards to an observer in a rowboat. 
The ``explanans'' comprises both general laws (refraction, water optical density) and antecedent conditions (an oar part in the water and part in the air, an oar consisting of a straight piece of wood). 
Hence, the ``explanandum'' is logically deduced from the ``explanans,'' thus the question ``Why does the phenomenon occur?'' is interpreted as ``What overarching principles and preceding circumstances lead to the phenomenon?'' 

Nevertheless, not all scientific laws are explainable through deduction, such as probabilistic or statistical ones. 
Thus, Hempel~\cite{hempel1962deductive} introduced a statistical systematization for scientific explanations, namely the Inductive-Statistical (I-S) model, recognizing the limitations of the Deductive-Nomological one. 
Hempel's I-S model is his natural way to extend the D-N model to statistical generalizations, remaining implicitly entrenched in the deductive ideal. 
Indeed, to explain means to express the probability of a given instance of $F$ being an occasion of $G$, represented by the variable $r$. 
Hempel's I-S explanation must be tied to all available reference knowledge, as stated by his ``maximal specificity'' requirement. 
The idea underlying this condition is the impossibility of genuine statistical explanation, that defines them as epistemically relative~\cite{Salmon1990}, and from which also Hempel derived the principle of ``high inductive probability,'' in which the value assigned to $r$ should be as close as possible to $1$~\cite{hempel2004aspects}. 

Explanation, according to these views, is the \emph{logical process} by which science provides answers to ``why questions'' and, thus, terms like ``comprehensible'' and ``understanding'' are considered to be inapplicable to scientific explanation since they do not fall within the domain of its logical aspects, to the extent that Hempel~\cite{hempel2004aspects} compared this process to the one of mathematical proofs. 
Future conceptions of explanation will increasingly focus on pragmatic aspects and probabilistic causality, moving further away from the deductive ideal. 

\subsection{Post-Hempelian Era}
After Hempel's ``received view'' a certain amount of formal and semi-formal models were proposed by different authors. 
Indeed, post-Hempelian scholars mainly rejected his conception of explanation and started from attacks on his model to build new interpretations.

\paragraph{Statistical Relevance Model.} Salmon~\cite{salmon1971statistical} moved from criticism about the inferential structure of explanation and proposed the Statistical-Relevance (S-R) model, which contemplates a specific idea of probabilistic causation. 
In his conception, explanations must consider not only events that respect the principle of ``high inductive probability'' but also unlikely ones. 
Statistical relevance determines to which homogeneous reference class the single event belongs. 
To establish homogeneity, the method involves partitioning non-homogeneous reference classes into maximal homogeneous sub-classes, which are mutually exclusive and comprehensive for the initial class. 
Thus, to explain means to place the ``explanandum'' in a chain of correlations expressed by statistical generalizations, that constitute the reference class meeting the maximal homogeneity criterion. 
A satisfactory theory of explanation should assign a fundamental role to causality, and, although statistical explanations are often discussed in seemingly indeterministic contexts, this does not negate the possibility of finding causal connections~\cite{Salmon1990}.

\paragraph{The Pragmatics of Explanation.} Van Fraassen~\cite{Pragmatics77VanFraassen}, unsatisfied with Salmon’s and previous accounts, introduced a pragmatic view of explanation. 
While the neo-positivist perspective was mainly concerned with establishing measures for verifying the validity of a scientific theory, such as its truthfulness or empirical adequacy, this view aims to determine the relevant part of a scientific fact by considering the contextual information, which relates to the knowledge and interests of the subject who posits the ``why question.'' Van Fraassen began by examining requests for specific ``why questions,'' which are comprised of a triplet $Q = \langle P_k, X, R\rangle$, namely, the topic, the antithesis class, and the relevance relation. 
The latter connects the informative part of the answer with the components of the question~\cite{van1980scientific}. 

\paragraph{The Unificationist View.} As the debate progresses, the importance of contextual elements in explanation increases. Friedman’s Unificationist view~\cite{friedman1974explanation} explored the feasibility of an objective conceptualization of scientific understanding, in seeking to clarify what is in the relationship between phenomena that determine one as the explanation of the other. 
The explanation process is not merely a substitution of one casual phenomenon. 
Rather, it involves replacing less comprehensive phenomena with more comprehensive ones, by reducing the number of independent events and enhancing our global understanding of the world. 
Indeed, unification is the element of the explanation relation that produces understanding. Kitcher~\cite{kitcher1981explanatory} proposed the most articulated Unificationist view, which posits that scientific activity aims to unify accepted knowledge, through general laws. 
Scientific understanding is achieved not by explaining individual occurrences, but by providing increasingly larger frameworks to fit them systematically. 

\paragraph{Abductive Explanation.}
The term ``abduction,'' often paralleled with the locution ``inference to the best explanation~\cite{Harman65},'' originated with Peirce~\cite{peirce1931collected}, who introduced it to signify a type of reasoning distinct from deduction, although not induction. Abduction is a type of nonmonotonic reasoning~\cite{Gottlob95} (i.e., defeasible inference) and consists of the process of forming explanatory hypotheses given a certain scenario~\cite{peirce1931collected}. 
The concept posits that when confronted with a phenomenon if one explanation emerges that plausibly accounts for the otherwise inexplicable, it is reasonable to lean towards accepting that explanation as likely correct~\cite{peirce1931collected}. 
After its first appearance, different formalizations have been suggested, taking the name of logic-based abduction, which is particularly suitable if complex causal relationships prevail~\cite{Gottlob95}.
However, the idea of inference to the best explanation is met with resistance in the field philosophy of science, as this kind of inference presupposes the truth of the explanatory premises~\cite{Salmon1990}. 
Indeed, what may be selected as the best explanation, could be within a group of incorrect ones~\cite{van1980scientific}. 
Moreover, this kind of explanation leaves open the role of pragmatic components for the selection of the~\textit{best} explanation for different individuals~\cite{van1980scientific}.

\paragraph{Neo-Mechanistic Theories.}
The Unificationist theory proposed by Kitcher~\cite{kitcher1981explanatory} sees explanation as global and, by referring to general laws, employs a top-down approach. 
On the other hand, causal-mechanical theories such as that advanced by Salmon~\cite{salmon1971statistical} employ a bottom-up approach and aim to describe the causal relationships involved in the phenomenon being explained~\cite{Salmon1990}. 
This type of explanatory knowledge seeks to provide understanding by showing the inner mechanism of phenomena of the world, that is, by exploring the internal workings of things, making it possible to open the ``black box'' of nature. 
During the '90s this account served as inspiration for neo-mechanistic theories, that proposed a more applicable view of causality aimed to identify mechanistic links~\cite{Salmon1990}, in a conception of causality understood as productivity.  
Among the most relevant, it is possible to encounter Glennan's~\cite{glennan_mechanisms_1996} Complex System account, in which
a mechanism consists of various behaviors comprising multiple components that can be separately analyzed and decomposed into smaller subsets. Additionally, the system's parts should exhibit a notable degree of robustness or stability. In other words, their properties should remain relatively constant in the absence of external interventions.  
A good explanation is made of an ``explanandum,'' which is the description of the phenomena to be explained, and an ``explanans,'' which is the inner mechanistic description.   
A different account comes from Bechtel and Abrahamsen~\cite{bechtel_explanation_2005}, which proposed the Decomposition and Localization model. 
Following their perspective, a mechanism is a structure that fulfills a function based on its constituent parts, its operations, and the overall organization.
Moreover, according to the authors, due to the epistemic character of explanations, representations, such as diagrams and verbal or linguistic descriptions, can support the inner mechanisms of nature.

\paragraph{Counterfactual Explanation.} In recent decades, a new type of approach to causality for explanation has gained popularity, namely the ``interventionist perspective''~\cite{woodward2004counterfactuals}. Specifically, an intervention is a perfected form of human experimental manipulation, devoid of anthropocentric components and described exclusively in terms of cause-and-effect and correlation~\cite{woodward2004counterfactuals}.
In the XAI literature, it is often argued that counterfactual knowledge can serve as a basis for causal understanding due to the contrastive nature of human explanation~\cite{miller2019explanation}, serving as the justification for laying the foundation for counterfactual models of explanation. 
However, terminological clarification is needed: counterfactuals and contrastive explanations are not synonymous, although they are often used as interchangeable terms~\cite{Paez19pragmatic}. 
The counterfactual explanation states that causal relations exist only if intervening on the cause $C$, produces a change in the effect $E$, remaining unchanged the relationship between the two variables~\cite{woodward2004making}. 
On the other hand, contrastive explanations answer the question ``Why x rather than y?'' instead of only ``Why x?''~\cite{Paez19pragmatic}.  
Nonetheless, the concept of counterfactual has a very wide and long tradition that goes beyond explanation. Indeed counterfactuals can be defined as conditional statements that discuss what would be the case if something were different~\cite{lewis2013counterfactuals}. This notion is closely related to that of possible worlds, denoting one of the differences between contrastive and counterfactual explanations: while the former can have a factual answer, the latter requires a hypothetical one~\cite{Paez19pragmatic}.

\section{A Comparison of Explainability Debates through an Epistemological Lens}
\label{sec:comparison}
Ultimately, we possess all the necessary tools to draw the analogy between scientific explanation and XAI debates, by looking at their pattern of development, as shown in Figure~\ref{fig:chrono}. 
As we noted, explanations were not initially accepted as distinct goals of science, since separate from description or prediction, in either realm. 
Indeed, in the domain of the philosophy of science, the acceptance of scientific explanations did not manifest uniformly from the origins of the debate, as various philosophers originally rejected the idea of considering them a distinct objective of science, favoring description~\cite{Salmon1990}. 
Over centuries, there has been a transition from discordant perspectives toward a major consensus, culminating in the proposal of diverse models for explanation. 
Analogously, within the discourse on XAI, explanations were not initially regarded as primary objectives of AI models, which predominantly sought predictive capabilities while prioritizing high accuracy~\cite{ali2023explainable}. This is also known as the interpretability/accuracy trade-off where the quest for improved predictive performance often comes at the cost of reduced model interpretability. This relationship has traditionally been viewed as mutually exclusive; however, this notion has been increasingly contested by several scholars that argue for optimization between both~\cite{ali2023explainable}.
Thus, it is possible and worth claiming, that the urgency for explanation did come after the need for accurate prediction and description in both the field of AI and the philosophy of science~\cite{ali2023explainable,Salmon1990}.

Moreover, we discerned a gradual change from logic-deductive models of explanation to statistical ones in both domains, as we witnessed a shift from certainty to uncertainty. 
Hempel's Deductive-Nomological model seeks explanations, by deducing from causal (or deterministic laws)~\cite{HempOpp48}. Additionally, in Hempel's~\cite{HempOpp48} first scripts, causal laws overlapped in their meaning with non-statistical laws, and although he recognized the existence of the latter, he restricted his account of explanation to the deductive ones. 
On the other hand, moving progressively forward in time, if we look into mechanistic or neo-mechanistic explanations, we encounter a progressive consideration of statistical relationships, while not losing the importance of causal connections. As Salmon states: 
\begin{displayquote}
    If indeterminism is true, some explanations will be irreducibly statistical—that is, they will be full-blooded explanations whose statistical character results not merely from limitations of our knowledge~\cite{Salmon1990}.
\end{displayquote}
As it is highlighted in Figure~\ref{fig:chrono}, if we move toward XAI, an interesting analogy emerges: the very first deductive expert systems, having rule-based knowledge, were directly interpretable and their explanation consisted of an inference of the output from the rules~\cite{clancey_fg_nodate}.
However, most ML models work as ``black boxes'' and their knowledge is opaque, so they don’t reveal sufficient details about their internal behavior~\cite{lipton2018mythos}. 
For this reason, differently from early rule-based systems, explainability in ML often seeks to find an interpretable model that approximates the original one, by finding statistical correlations~\cite{ding2022explainability} (e.g., many explainability methods offer summary statistics for each feature, such as feature importance~\cite{ali2023explainable}). 
However, genuine causal relationships must be preserved~\cite{Salmon1990}. 
Moreover, manipulative-counterfactual approaches to explanation have increasingly gained popularity in both debates. 
On the side of scientific explanation, by advancing an intervention-centered notion of causality~\cite{woodward2004counterfactuals}. While, on the other of AI, showing what should have been different to change the decision of the system. Specifically, consisting of the smallest change that can be made to a particular instance to get a different decision from the AI~\cite{wachter2018counterfactual}.   

Lastly, a typical distinction found in XAI literature is within the categorization of global and local explanations, the first ones aimed to explain the knowledge of general patterns of the system as a whole, while the latter, a single decision~\cite{doshi2017towards}. 
As underlined in Figure~\ref{fig:chrono}, scientific explanation, in a broader sense, sees patterns of explanation underlying a similar distinction between top-down and bottom-up accounts. 
The first one is, in this sense, global, as it relates to the structure of the whole world~\cite{Salmon1990}. 
The second one, as can be seen very well in Bechtel and Abrahamsen's account~\cite{bechtel_explanation_2005}, aims to identify the relationships and explanations of individual parts. 

\subsection{Going Deeper: Concepts of Explainability}
\label{sec:deeper}
In addition to establishing connections between the two debates as a whole, it is possible to examine analogies between related concepts and common terminology that we identified in our comparison of scientific explanation and XAI. 
This analysis considers preliminary epistemological implications that are relevant within the XAI domain, such as the relationship between explanation and understanding, the significance of similarity in explanation, and the desiderata of good explanations, thereby laying foundational groundwork for future research.

\paragraph{The Epistemological Relation between Explaining and Understanding.}
The earliest theories of scientific explanation, proposed by the analytical philosophical tradition, were not concerned with understanding, as they claimed that it was not part of the explanation relation. 
According to Hempel~\cite{HempOpp48}, a scientific explanation is restricted to deductive and logical inference, by which science answers ``why questions'' and, thus, he considered terms like ``comprehensible'' and ``understanding'' out of its domain~\cite{hempel2004aspects}. 
However, with the evolution of the debate, pragmatic factors have been taken into consideration increasingly, appearing awareness of the fact that an explanation should be considered with reference to a specific question~\cite{Pragmatics77VanFraassen}. 
Hence, an explanation is not decontextualized but pertains to the situation in which questions and answers are posed.
On the other hand, the XAI field has started to progressively consider the importance of a diverse pool of users and different stakeholders when providing explanations~\cite{ali2023explainable,dwivedi2023explainable}, determining the appearance of terms such as ``interpretability,'' and ``understandability,'' around the XAI context~\cite{lipton2018mythos,Paez19pragmatic}.
In general terms, explanations involve~\textit{understanding} how the world works. 
However, the epistemic relation between explanation and understanding is not straightforward~\cite{Salmon1990}. 
In the context of XAI, this implies that a prior grasp of what it signifies that a subject understands a model or a decision is required~\cite{Paez19pragmatic}. 
However, philosophers have engaged in extensive reflections which can suggest how to delineate the precise factors that contribute to the generation of understanding. 
For instance, notwithstanding not lingering on understanding, Hempel~\cite{HempOpp48} posits that it consists of seeing the phenomenon in question as an instance of a general pattern. 
Furthermore, Friedman's Unificationist view~\cite{friedman1974explanation} claims that science increases understanding by reducing the total number of independent phenomena. To wit, the phenomenon to be explained is replaced with a more comprehensive one, reducing the total number of phenomena. Finally, Salmon~\cite{Salmon1990} asserts that explanations seek to provide a systematic understanding of empirical phenomena by showing how they fit into a causal nexus. 

\paragraph{Similarity, Familiarity, and Surrogate Models.}
Explanation of ML often consists of adopting a surrogate and interpretable model, such as linear regression, that should provide representations necessary to obtain understanding~\cite{ding2022explainability}. 
However, a relevant issue is establishing why this surrogate serves as an explanation of the original model. 
Indeed, for any XAI model, there should be a formal linkage, such as isomorphism or similarity, between it and the initial model~\cite{duran2021dissecting}. Nevertheless, the majority of surrogate models used currently lack rigorous assurances, raising uncertainties about the efficacy of these approximations in elucidating decision-making processes~\cite{marques2023explainability}.
On the other hand, formal explanations seek to establish guarantees or justifications with respect to the determined explanation~\cite{Audemard2021TradingCF,Izza021}, such as Random Forest explanations with SAT~\cite{Izza021} or abductive explanations~\cite{Amogud21Abductive}. 
Without this type of connection, there is no basis to state that an explanation provided by the XAI model applies to a ``black box''~\cite{duran2021dissecting,marques2023explainability}. Similarly, some philosophers of science have argued that understanding can given by familiarity, in which the ``explanans'' is an approximation similar to the ``explanandum'' or an idealization of it~\cite{van1980scientific}. 
However, to others this view is deemed inadequate: being familiar gives no grounds for being understood and, regardless some explanations might evoke a feeling of familiarity, this is not a relevant factor in sound explanations~\cite{friedman1974explanation,HempOpp48}.

\paragraph{\textit{Bona Fide} Explanations Criteria.} 
Also as a consequence of the aforementioned considerations, researchers in both epistemology and the XAI domains have sought to identify the characteristics that distinguish~\textit{bona fide} explanations, i.e., explanations should satisfy certain requirements to be considered valid~\cite{carnap1962logical,duran2021dissecting,Salmon1990}.
For instance, Miller has done incipient work in establishing criteria to evaluate XAI, by deriving principles from social sciences~\cite{miller2019explanation}. 
Moreover, Mueller et al.~\cite{mueller2021principles} provided an exhaustive list of principles that emerged within XAI literature. 
Within the epistemological domain, Hempel~\cite{hempel1962deductive,HempOpp48} introduced the principle of factuality, namely that the ``explanans'' and the ``explanandum'' must be true. 
Conversely, a potential explanation possesses all the essential characteristics of a valid explanation, except for the truth~\cite{hempel1962deductive}. 
Carnap~\cite{carnap1962logical} identified four criteria for explanations: similarity to the ``explicandum,'' exactness, fruitfulness, and simplicity. 
Specifically, similarity to the ``explicandum'' refers to the necessity of the ``explicatum'' to adequately correspond to the ``explicandum,'' otherwise, it fails to fulfill the intended function of the concept it is meant to substitute. 
Exactness denotes that explanations replace a less precise concept with a more precise one. 
Fruitfulness reflects the fact that the ``explicatum'' should offer profound insights. 
Simplicity states that the ``explicatum'' should be as simple as the previous requisites allow.
The profusion of criteria derived from epistemology and the proximity of the two domains suggest that epistemological principles may also serve as a source of inspiration in evaluating what makes a good explanation, helping in the assessment of theoretical guidelines for evaluating XAI derived from the philosophy of science.

\section{Conclusions}
The concept of explainability has been the object of numerous inquiries. 
However, notwithstanding its acknowledgment as a fundamental right and the considerable number of proposed models, it is widely criticized for not having convincing and unifying conceptual grounds. 
This article tries to fill in this gap and aims to contribute to the foundations for the construction of a ``bridge'' between epistemology and ML, which may lead to deeper explorations of epistemological consequences of AI explanations. 
We compared two apparently different debates, scientific explanation, and XAI, in an attempt to assist XAI discussion with a well-grounded philosophical foundation. 
We traced the history of their development, criticisms that have emerged, and key concepts, examined through the epistemological lens. An intriguing picture has emerged:~\textit{the development of the debates followed a general common progression, specifically from deductive to statistical explanations.} 
Interestingly, we also notice that similar concepts have independently arisen in both realms, such as the relation between explanation and understanding, the importance of pragmatic factors, the relationship between similarity and explanation, and the search for~\textit{bona fide} explanations. 
Hence, in Section~\ref{sec:comparison} we have briefly illustrated how possible implications can be derived from epistemology in order to analyze XAI concepts.
We identified the roots from which philosophical terminology has originated and also of a ``dictionary'' of shared concepts, to help XAI practitioners draw insights from past philosophical debates and their implications. 
Hence, future work may be aided by the instruments of the philosophers that we hope to have enlightened. 
Moreover, we offer ML researchers extensive epistemological literature, from which they can draw inspiration. For example, counterfactual explanations, with their deep roots in philosophy, have recently garnered attention in the field of XAI, demonstrating practical utility across various applications. 
Similarly, we aim to propose novel ideas to inspire further research. 
Our work can be seen as a thoughtful philosophical guide based on a comparative analysis of two pieces of literature that have been little explored in their synergy, however so close to each other.

\section*{Acknowledgment}

This work has been partially supported by: (i)  EU - NGEU National Sustainable Mobility Center (CN00000023) Italian Ministry of University and Research Decree n. 1033—17/06/2022 (Spoke 10); (ii) project SERICS (PE00000014) under the NRRP MUR program funded by the EU - NGEU.


\bibliographystyle{ieeetr}
\bibliography{main}

\end{document}